\documentclass[11pt,a4paper]{article}
\usepackage[hyperref]{acl2020}
\usepackage{times}
\usepackage{latexsym}
\usepackage{amsmath}
\usepackage{graphicx}
\usepackage{amsfonts}
\usepackage{url}
\usepackage[ruled,vlined]{algorithm2e}

\usepackage{microtype}

\aclfinalcopy 

\title{Probabilistically Masked Language Model Capable of \\Autoregressive Generation in Arbitrary Word Order}

\author{Yi Liao, Xin Jiang, Qun Liu \\
  Huawei Noah's Ark Lab\\
  \texttt{\{liaoyi9, jiang.xin, qun.liu\}@huawei.com}\\}

\date{}

\begin{document}
\maketitle
\begin{abstract}
Masked language model and autoregressive language model are two types of language models. While pretrained masked language models such as BERT \cite{bert} overwhelm the line of natural language understanding (NLU) tasks, autoregressive language models such as GPT \cite{gpt} are especially capable in natural language generation (NLG). In this paper, we propose a probabilistic masking scheme for the masked language model, which we call \textit{probabilistically masked} language model (PMLM). We implement a specific PMLM with a uniform prior distribution on the masking ratio named u-PMLM. We prove that u-PMLM is equivalent to an autoregressive permutated language model. One main advantage of the model is that it supports text generation in arbitrary order with surprisingly good quality, which could potentially enable new applications over traditional unidirectional generation. Besides, the pretrained u-PMLM also outperforms BERT on a set of downstream NLU tasks.
\end{abstract}
\section{Introduction}
Large-scale pretrained language models \cite{t5,alice,albert,roberta,tinybert} have drawn lots of research attention as these models have brought significant improvements to many NLU and NLG tasks. As a major category of pretrained language models, masked language model (MLM) \cite{bert, spanbert} is trained using a denoising autoencoding objective. In a typical MLM, some tokens in a sentence are replaced by a special token \texttt{[MASK]}. The training objective is to predict the original tokens that are masked in the sentence. As the first large-scale pretrained masked language model, BERT chooses to mask 15\% of the tokens in sentences randomly. Following BERT, various language models have been proposed with different masking schemes.

\begin{figure}[t]
    \begin{center}
        \noindent\fbox{%
            
            \parbox{0.4\textwidth}{%
                \textbf{The} wolf has an extraordinary speed , and it can often jump from a spot \textbf{quick} enough to escape a spot already occupied by an adult wolf . Unlike the \textbf{brown} and black bear , where it is easily distracted by wolves , the gray \textbf{fox} does not run over a wolf , and is often driven mad . Having \textbf{jumps} with high speed that breaks the wolf ' s legs before it is run \textbf{over} , a grey wolf could defend itself against an adult of other species as \textbf{the} best predator at any time . The black bear may kill packs of four \textbf{lazy} , though the gray fox can inflict significant wounds on a \textbf{dog} .
            }%
        }
    \caption{A piece of text generated by a PMLM in random order. The bolded words, which compose the input sentence ``The quick brown fox jumps over the lazy dog", are distributed across the paragraph with a predefined length. The blank spaces are filled by the model in a random order to form the complete paragraph.}
    \label{intro-example}
    \end{center}
    \vspace{-10pt}
\end{figure}

While the pretrained masked language models achieve state-of-the-art performances in a line of downstream NLU tasks, researchers pay more attention to autoregressive language model when it comes to text generation. Unlike predicting the masked tokens, the autoregressive language model learns a sequential generative process of text sequences. Hence it naturally performs better for natural language generation. For example, GPT-2 \cite{gpt-2} as well as Transformer-XL \cite{transformerxl}, is able to generate fluent and coherent paragraphs of text that highly resembles human writings.

\begin{figure*}
    \centering
    \includegraphics[scale= 0.7]{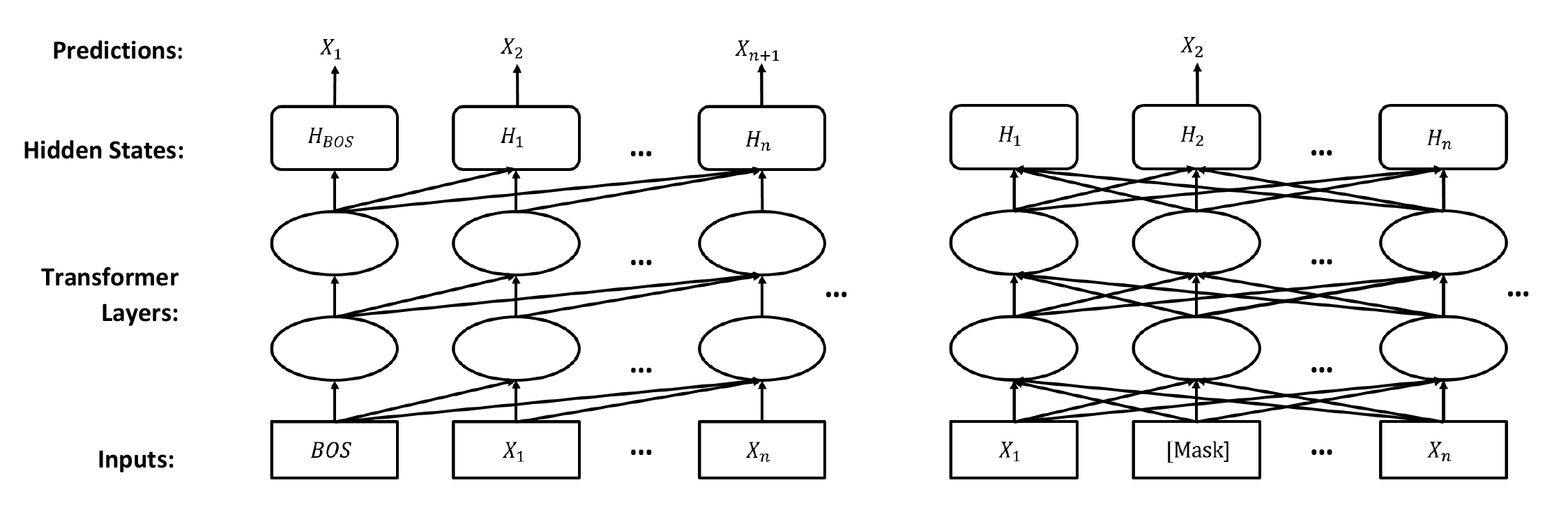}
    \caption{The structures of autoregressive language model (left) and masked language model (right).}
    \label{fig:lm}
\end{figure*}

In this paper, we propose a \textit{probabilistically masked} language model (PMLM) to bridge the gap between masked and autoregressive language models. The basic idea behind the connection of two categories of models is similar to MADE \cite{germain2015made}. PMLM is a masked language model with a probabilistic masking scheme, which defines the way sequences are masked by following a probabilistic distribution. While the existing work proposes masking strategies aiming at improving the NLU abilities, PMLM addresses the generation capability in particular. Besides, as a masked language model, PMLM maintains its strong ability in natural language understanding.

In addition to the traditional unidirectional (e.g., left-to-right) generation, a unique ability for PMLM is to autoregressively generate sequences \textit{in arbitrary order}, and the generated sequences are still of high quality. In contrast to traditional left-to-right generation, arbitrarily ordered text generation has two main characteristics. First, the next token to be predicted could be in any position that is masked. Second, the next token to be predicted depends on all the previous observed/generated tokens. Arbitrarily ordered generation enables more interesting applications than unidirectional generation. For example, Figure \ref{intro-example} shows an example of \textit{cloze test}, where the prompted text ``The quick brown fox jumps over the lazy dog'' is distributed across a paragraph with a predefined length, and the task is to predict all the surrounding words and complete the paragraph.  
This is actually very challenging for conventional generation models since when predicting each word, the fluency and coherence of text are hard to be guaranteed given the contextual constraints on both sides. More applications may include acrostic poetry generation, news generation based on given facts, machine translation with lexical constraints, etc.

We employ a simple uniform distribution of the masking ratio and name the model as u-PMLM. We prove that u-PMLM actually learns an autoregressive language model on random permutations of training sequences. The experiments show that the quality of text generated by u-PMLM in arbitrary order is as good as that generated by GPT in sequential order. Besides, u-PMLM outperforms BERT significantly on the GLUE benchmark for natural language understanding.

\section{Preliminary}
\subsection{Transformer}
Transformer \cite{attention} is the backbone model for many pretrained language models. Transformer is composed of a stack of multi-head self-attention and token-wise feed-forward layers. At each layer, the hidden state of each token is updated based on the historical hidden states computed in the lower layer. Let $X = \{x_1, x_2, ..., x_N\}$ denote the sequence of tokens, where $N$ is the length of the sequence. Fed with $X$ as input, the final output of the Transformer, denoted as $H = \{h_1, h_2, ..., h_N\}$, captures the contextual representation of the tokens in the sequence.

\subsection{Autoregressive Language Model}

In autoregressive language model, the sequence generation process is modeled as a Markov chain, where the token to be predicted depends on all the previous tokens. The training objective can be formulated as:
\begin{equation} \label{alm}
L_{\mathrm{alm}}(X) = \sum_{n=1}^{N}\log p(x_n|x_1, ..., x_{n-1};\theta),
\end{equation}
where $\theta$ denotes the parameters of the model. Figure \ref{fig:lm}(a) shows the diagram of autoregressive LM. In the model, the $n$-th token can only attend on the tokens at positions less than $n$. The autoregressive model is usually trained in the way of \textit{teacher-forcing}, i.e., always using the ground-truth tokens as inputs and outputs in training.

Pretrained autoregressive models such as GPT \cite{gpt, gpt-2} are especially capable of generating fluent and coherent text that highly resembles human-written text. However, unidirectional attention brings two limitations. Firstly, autoregressive model as in Figure \ref{fig:lm}(a) can only generate text from left to right; Secondly, unidirectional attention blocks the contextual information from the right side of the current token, affecting the completeness of the contextual representation.

\subsection{Masked Language Model}
To obtain complete representations of the tokens in a sequence, researchers resort to bidirectional attention as shown in Figure \ref{fig:lm}(b). 
Specifically, the training instances are created by replacing a subset of tokens in the input $X$ with a special token \texttt{[MASK]}, and the objective is to predict the masked tokens. Such model is called masked language model (MLM). Let $\Pi = \{\pi_1,\pi_2,..., \pi_K\}$ denote the indexes of the masked tokens in the sentence $X$, where $K$ is the number of masked tokens. Let $X_{\Pi}$ denote the set of masked tokens in $X$, and $X_{-\Pi}$ denote the set of observed (unmasked) tokens. The objective of MLM is:
\begin{equation}\label{mlm}
\begin{split}
L_{\mathrm{mlm}}(X_{\Pi}|X_{-\Pi}) = \frac{1}{K}\sum_{k = 1}^K \log p(x_{\pi_k}|X_{-\Pi};\theta).
\end{split}
\end{equation}
The assumption in Equation \ref{mlm} is that the probability of predicting a masked token is independent of each other. BERT \cite{bert} is a typical masked language model.

Due to the incorporation of bidirectional attention, masked language model can capture the contextual information on both sides. Consequently, it usually achieves better performances when finetuned in downstream NLU tasks than the conventional autoregressive models. However, the masking scheme and the independence assumption also affect its performance on text generation compared to autoregressive models \cite{bertmouth}.

\section{Probabilistically Masked Language Model}
Different masking schemes have been proposed for pretraining the masked language model. The most straightforward masking scheme is to randomly mask tokens in sentences in a fixed ratio, e.g., 15\% in BERT. Following BERT, various models have proposed modifying the masking scheme to improve its NLU capability. ERNIE \cite{ernie} proposes the entity-level masking and phrase-level masking, where the words composing an entity or phrase are masked as a whole. SpanBERT \cite{spanbert} proposes to mask a continuous random span of text rather than random tokens. These masking strategies have shown to be effective for certain classes of NLU tasks. 

In contrast to the existing work, we propose a \textit{probabilistic masking} scheme that tries to improve the text generation ability of the masked language model. Probabilistically masked language mode (PMLM) is a natural generalization of the MLM with a probabilistic masking ratio. It assumes that the masking ratio is drawn from a probabilistic distribution. Therefore, each training instance is associated with a different masking ratio sampled from the given distribution.

    
%
\subsection{Model Formulation}
To give a formal definition of the PMLM, we need to elaborate the training objective defined in Equation \ref{mlm}. Let $M = \{m_1, m_2, ..., m_N\}$ denote a sequence of binary variables indicating which token in $X= \{x_1,x_2,...,x_N\}$ is masked. $m_n = 1$ indicates $x_n$ is masked, and $m_n=0$ otherwise. Noted that since $\Pi = \{\pi_1, \pi_2, ..., \pi_K\}$ denotes the indexes of masked tokens, $m_{\pi_k} = 1$ holds for any $\pi_k \in \Pi$. Considering $M$ as latent variables, the expected log-likelihood function of observing $X_{\Pi}$ conditioning on $X_{-\Pi}$ over all possible $M$ is:
\begin{equation}\label{PMLM_latent}
\begin{split}
L_{\mathrm{pmlm}}(&X_{\Pi}|X;\theta) \\
=& \mathbb{E}_{M|X} [\log p(X_{\Pi}|X_{-\Pi})]\\
=& \sum_{M} [\log p(X_{\Pi}|X_{-\Pi}; \theta)]p(M|X)\\
\end{split}
\end{equation}
The term $\log p(X_{\Pi}|X_{-\Pi}; \theta)$ is identical to the objective function in Equation \ref{mlm} for a deterministic mask $M$.
In the vanilla MLM, it is assumed that $M$ are i.i.d. for each position and independent to $X$, namely,
\begin{equation}
p(M|X) = p(M) = r^K(1-r)^{N-K},
\end{equation}
where $r$ is the masking ratio. 

Most existing MLMs such as BERT simply set a fixed value to the masking ratio $r$. In our proposed PMLM, however, we assume $r$ is a random variable drawn from a prior distribution $p(r)$. Therefore, the distribution $p(M)$ becomes:
\begin{equation}
\begin{split}
    p(M) = \alpha_M = & \int p(M|r)p(r)dr \\
         =& \int r^K(1-r)^{N-K} p(r) dr \\
\end{split}
\end{equation}

With above derivations, we can formulate the expected log-likelihood function of PMLM as:
\begin{equation}\label{PMLM_X}
\begin{split}
&L_{\mathrm{pmlm}}(X_{\Pi}|X;\theta) \\
&=\sum_{M} [\log p(X_{\Pi}|X_{-\Pi};\theta)]\alpha_M\\
&=\sum_M \frac{\alpha_M}{K} \sum_{k = 1}^K  \log p(x_{\pi_k}|X_{-\Pi};\theta)
\end{split}
\end{equation}

Equation \ref{PMLM_X} is optimized by sampling $M$ according to the prior distribution over the training set. By controlling the prior distribution, we can cover a wider range of sequence prediction tasks in training, which can potentially enhance the representation power of the pretrained model. For instance, in the left-to-right autoregressive model, the masking ratio is uniformly distributed across different positions, which makes the model learn to generate the next token given the previous context of different lengths. This inspires us to try the uniform prior on masking ratio for PMLM.

\subsection{PMLM with a uniform prior}\label{sec:u-PMLM}
u-PMLM is an implementation of PMLM with a continuous uniform distribution on the masking ratio: 
\begin{equation}\label{uniform}
p(r)=\left\{
    \begin{aligned}
    1&, 0 \leq r\leq 1 \\
    0&, \text{otherwise.}
    \end{aligned}
    \right.
\end{equation}
Like most pretrained language models, the backbone model for u-PMLM is Transformer as well. 

We prove that u-PMLM is equivalent to the autoregressive permutated language model (APLM) by recombination of the factorized log-likelihood function, which is basically the autoregressive language model trained on all possible permutations of the training instances:
\begin{equation} \label{aplm}
\small
L_{\mathrm{aplm}}(X) = \mathbb{E}_{\sigma} \left[\sum_{t=1}^{N}\log p(x_{\sigma_t}|x_{\sigma_1},\ldots,x_{\sigma_{t-1}};\theta) \right],
\end{equation}
where $\sigma$ denote random permutations. The detail derivation is included in the Appendix \ref{proof}.

Ordinary autoregressive model can be regarded as a special case of the permutated model.
Therefore, we can expect that the u-PMLM is able to work as the autoregressive model in sequential prediction. Moreover, since it can handle any permutation of the sequence, it should have the ability to generate sequences in arbitrary word order.

\subsection{Generation with u-PMLM}
Algorithm \ref{algorithm} depicts the algorithm to autoregressively generate a sequence in random order with u-PMLM. The process starts with a sequence containing full of the special token \texttt{[MASK]}. Then the model iteratively replaces a \texttt{[MASK]} token in a random position with a predicted token, until all the tokens are predicted. An example showing the states of the sequence during the generation process is presented in Table \ref{random_generation}. The generation order could be arbitrary, which is much more flexible than the traditional unidirectional generation. On the other hand, our model can not automatically determine a best generation order, which could be a interesting problem for future research.
\begin{algorithm}[tbh]
    \KwResult{Generated Text Sequence $S = \{s_1, s_2,...,s_N\}$}.
    \textbf{Initialization}: \\
    i. A sequence $S$ with all \texttt{[MASK]} tokens.\\
    ii. Unvisited index set $U = \{1,2,...,N\}$.\\
    \While{$U$ is not empty}{
        1. Randomly pick a number $n$ from $U$;\\
        2. Input u-PMLM with $S$ and predict the $n$-th token $x_n$; \\
        3. Replace the $n$-th token of $S$ with the predicted token $x_n$, i.e., $S(n) \leftarrow x_n$;\\
        4. Remove $n$ from $U$.
    }
    \caption{Generation with u-PMLM}
    \label{algorithm}
\end{algorithm}

\begin{table*}
    \centering
    \begin{tabular}{c|c|cccccccc}\hline
        \textbf{Step}&\textbf{Prediction Index}&\multicolumn{8}{c}{\textbf{State of the sequence}}\\ \hline
        0& n/a&\_&\_&\_&\_&\_&\_&\_&\_\\    
        1&3 &\_&\_&a&\_&\_&\_&\_&\_\\
        2&7     &\_&\_&a&\_&\_&\_&random&\_\\        
        3&1&This&\_&a&\_&\_&\_&random&\_\\    
        4&2&This&is&a&\_&\_&\_&random&\_\\    
        
        5&4&This&is&a&sentence&\_&\_&random&\_\\    
        6&6&This&is&a&sentence&\_&in&random&\_\\    
        7&5&This&is&a&sentence&generated&in&random&\_\\    
        8&8 &This & is & a & sentence & generated & in & random & order\\ \hline
        \multicolumn{9}{l}{Generation Order: 3$\rightarrow$7$\rightarrow$1$\rightarrow$2$\rightarrow$4$\rightarrow$6$\rightarrow$5$\rightarrow$8}\\ 
        \multicolumn{10}{l}{Output: This is a sentence generated in random order} \\ \hline
    \end{tabular}
    \caption{An example of how u-PMLM generates a sequence in random order. The special token \texttt{[MASK]} is simplified as the symbol ``\_".}
    \label{random_generation}
\end{table*}

\paragraph{Positional Embedding}
Most pretrained masked language models have employed absolute positional embedding to incorporate the positional information of the input tokens. We train two variants for u-PMLM, one with absolute positional embedding and the other with relative positional embedding \cite{rpe}. The experiments show that NLG ability is not sensitive to relative or absolute positional embedding, while NLU ability is improved with relative positional embeddings.

\paragraph{Model Inference}
Although both u-PMLM and GPT generate sequences autoregressively based on Transformer, they are slightly different at inference time. For u-PMLM, since we use the bidirectional Transformer, each time a token is generated, the hidden states of all the tokens need an update. For GPT, since the unidirectional Transformer is employed, the latter generated token does not affect the hidden states of previous tokens. This can result in different computational complexity. However, since a typical Graphics Processing Unit (GPU) computes matrices in parallel, the actual difference in inference time is not that significant. We report the comparison of time consumption in the experimental section.

\subsection{Training Settings}\label{sec:training-setting}
\paragraph{Model Size}: The size of our pretrained u-PMLM is identical to BERT-base, which contains 12 hidden layers and 12 attention heads. The hidden size is 768, and the intermediate size is 3072. The dropout rate is set to 0.1.

\paragraph{Training Data}
We employ the commonly adopted training data, namely BookCorpus and Wikipedia to train our u-PMLM model. We obtain 4.1 Gb for the BookCorpus dataset and 11.9 GB for the Wikipedia dataset after data cleaning. We further employ the same vocabulary and tokenization techniques as BERT for converting the text sequences to ID sequences. The vocabulary contains 28,996 cased tokens. We set the maximum sequence length to 128.
\paragraph{Training Platform}
We train u-PMLM using Horovod framework with 56 NVIDIA V100 (32GB) GPUs. To speed up the training process, we employ mix-precision training technique. The batch size is set to 150 for every single GPU, thus the total batch size is 8400. The optimizer is Lamb Optimizer~\cite{you2019large}, which is more suitable for large batch size than Adam Optimizer. We train u-PMLM for 600K steps, taking roughly 135 hours in total.

\section{Experiments}
We evaluate both the natural language generation ability and natural language understanding ability of u-PMLM trained in the settings described in Section \ref{sec:training-setting}. 
\subsection{Comparative Models}
We train the BERT model and GPT model as the comparative models in the experiments. BERT and GPT are representative models for masked language model and autoregressive language model, respectively.
To make fair comparisons, we train both models from scratch using the same settings described in Section \ref{sec:training-setting}, including the same training platform, model size, training data, vocabulary, and training steps. Note that since BERT adopts absolute positional embedding, the variant for u-PMLM with absolute positional embedding is trained for a fair comparison with BERT. Throughout the experimental section, u-PMLM-R and u-PMLM-A are short for the variants with relative and absolute positional embeddings, respectively.
\subsection{Autoregressive Generation}
\paragraph{Perplexity Evaluation}
Perplexity (PPL) measures the quality of a language model, where the task is to predict the next word or character in a document. Typically, the predicting order follows the generation order. However, as bidirectional u-PMLM and BERT supports text generation in arbitrary order. Hence we also evaluate the perplexity when predicting words in arbitrary order.

We evaluate the perplexity using two datasets for evaluating perplexity. The first dataset, Wikitext103, is a collection of over 100 million tokens extracted from the set of verified Good and Featured articles on Wikipedia. The second dataset, One-Billion Words, consists of 829 million tokens derived from a news-commentary site. Both datasets are widely adopted for evaluating language models. However, there are significant differences between these two datasets in terms of the length of sequences. The Wikitext103 dataset is more similar to the pretraining datasets, containing long articles. On the other hand, the One-Billion Words dataset contains only single sentences, roughly half of which contain less than 24 tokens. We have ensured that all the three models have the same context length, the same vocabulary, as well as the same tokenization method, which would affect the perplexity values. For Wikitext103 dataset, the context length is set to 128, and each context containing multiple coherent sentences. For the One-Billion Words dataset, context length is set to 50. Short sentences are appended with \texttt{[PAD]} to reach length 50. Actually, the context for nearly all the sentences is shorter than 50. Both datasets provide training and test sets. We first finetune the model using the training set before evaluating perplexity on the test set. For each model, the algorithm for the finetune phase is the same as that for the pretraining phase.

\begin{table}[t]
    \begin{tabular}{|l|l|l|}
        \hline
        Model            & PPL(sequential) & PPL(random) \\ \hline
        BERT & 23.12          & 25.54       \\ \hline
        GPT       & 21.23        &N/A             \\ \hline
        u-PMLM-R   & 19.58          & 21.51       \\ \hline
        u-PMLM-A   & 19.32          & 21.30       \\ \hline
    \end{tabular}
\caption{Perplexity on Wikitext103.}
\label{ppl-wikitext103}
\end{table}

\begin{table}[t]
    \begin{tabular}{|l|l|l|}
        \hline
        Model            & PPL(sequential) & PPL(random) \\ \hline
        BERT &        140.67   &   56.97     \\ \hline
        GPT       & 24.25       &    N/A         \\ \hline
        u-PMLM-R   &    35.24   &    38.45    \\ \hline
        u-PMLM-A   &       49.32    &   42.46     \\ \hline
    \end{tabular}
    \caption{Perplexity on One-Billion Words.}
    \label{ppl-1billion}
\end{table}
The evaluation results of perplexity are shown in Table \ref{ppl-wikitext103} and Table \ref{ppl-1billion}. ``Sequential" refers to the traditional left-to-right text generation, while for ``random", each sentence in the test set is assigned a random generation order. Smaller PPL indicates better language model performance. We first investigate the performance on Wikitext103 dataset. We observe that the PPL for u-PMLM is comparable to GPT on Wikitext103 dataset, indicating that the language model learned by u-PMLM is as good as GPT when the context length is sufficiently long. In such case, the text generated by u-PMLM is as good as GPT.  Moreover, the PPL of u-PMLM for randomly ordered language model is comparable to the left-to-right generation, which implies that u-PMLM has a strong ability for arbitrarily ordered generation. Besides, the results show that there are few differences between relative positional embedding and absolute positional embedding for u-PMLM. On the other hand, although BERT supports generation in arbitrary word order as well, the PPL for BERT is significantly worse than our proposed u-PMLM for both ``sequential" and ``random" settings, demonstrating the effectiveness of the proposed probabilistic masking scheme. We show more cases of text generation in random order for u-PMLM-A and BERT in Appendix \ref{text-gen-upmlm-bert}.

However, for PPL on One-Billion Words, the performances of u-PMLM and BERT are not satisfactory in comparison with GPT. Generally, PPL for all these models increases on One-Billion Words dataset as the context length becomes much smaller, which also reflects PPL's relationship to context length. The reason might be the large portions of \texttt{[PAD]} in the One-Billion Words dataset, i.e., more than 50\% of the context for nearly 50\% of the training instances are filled by \texttt{[PAD]}. We suspect that the \texttt{[PAD]}s affect the prediction process for bidirectional models.  On the other hand, unidirectional models such as GPT naturally ignore the effect of \texttt{[PAD]} tokens in the tail of context. The results imply that u-PMLM could be further improved in the future to be more robust.

\paragraph{Latency}
As analyzed in Section \ref{tab:time-complexity}, the time complexity for generation for masked language model is $N$ times of autoregressive language model when computing the hidden states in each Transformer layer. However, when employed for text generation on GPU, the difference might be less significant. We test the latency for generating 100 128-length sentences for GPT and u-PMLM respectively. The computational platform is NVIDIA V100 GPU. The results are shown in Table \ref{tab:time-complexity}. The results show that u-PMLM costs roughly 20.1\% more time than GPT for generating sentences, which is much less than the theoretical time complexity difference.
\begin{table}[t]
    \centering
    \begin{tabular}{|l|l|} \hline
         Models& Cost Time \\ \hline
         GPT&        105.6 s     \\
         u-PMLM-A &   126.8 s  \\ \hline
    \end{tabular}
    \caption{Latency for generating 100 128-length sequences.}
    \label{tab:time-complexity}
\end{table}
\paragraph{Comparison With GPT for Generation}
In the introduction section, we have shown an example showing the application of arbitrarily ordered text generation, where the tokens in the input sentences are distributed across the generated sentences. Indeed, the major difference with GPT is that the input text could be inserted anywhere in the generated text, which makes the generation process more controllable. Meanwhile, the output text contains certain predefined tokens.

Figure \ref{case1} and Figure \ref{case2} shows the generated paragraphs of GPT and u-PMLM, respectively. For GPT, the input text can only be placed in the beginning and the generation process become uncontrollable, resulting in generating sentences with topic drift. In contrast, u-PMLM allows manually placing anchor sentences in the middle or end of the generated text to guide the topic of the generated text. As shown in Figure \ref{case2}, we place  ``Tom is a cat and Jerry is a mouse ." and ``Tom and Jerry become good friends in the end ." at the beginning and end of the paragraph. The middle parts are generated by u-PMLM from left-to-right. Such generation method allows us to better retain the topic of the generated content.

\begin{figure}[t]
    \begin{center}
        \noindent\fbox{%
            
            \parbox{0.4\textwidth}{%
                \textbf{Tom is a cat and Jerry is a mouse .}`` It ' s very sad ! " . The writers had wanted Tom to have `` something big to tell it . . . and a fun place to get excited " . The writers believed that the `` little animal " and the `` little black dog " at the end of the episode would have attracted more attention from viewers , but it never took place . Tom ' s first television role was that of the boy scout `` Mr . Krabs " in the 1978 NBC Western comedy pilot , The Search for Mr . Krabs .
            }%
        }
    \caption{Unidirectional Text Generation with GPT}
    \label{case1}
    \end{center}
    \vspace{-10pt}
\end{figure}

\begin{figure}[t]
    \begin{center}
        \noindent\fbox{%
            
            \parbox{0.4\textwidth}{%
                \textbf{Tom is a cat and Jerry is a mouse .} However , the two have a common . The first part is a joke about Jerry and Tom fighting in the middle of the episode . The two get on the run from the restaurant , and Tom ' s mother is shocked that they would have to do so . After a few minutes , Jerry arrives and decides to have a fight . The two go to the casino , where Jerry tries to fight them back by using a splint of grease and a bucket of wine in the bar . They reunite at a restaurant dance , and \textbf{Tom and Jerry become good friends in the end .}
            }%
        }
    \caption{Bidirectional Text Generation with u-PMLM }
    \label{case2}
    \end{center}
    \vspace{-10pt}
\end{figure}

\subsection{Natural Language Understanding}
\begin{table*}
 \small
    \begin{tabular}{|l|l|l|l|l|l|l|l|l|l|}
        \hline
        Model        & COLA & SST2 & MRPC      & STSB      & QQP       & MNLI-m/mm & QNLI & RTE   & AVG. \\ \hline
        BERT(A)    & 52.1 & 93.5 & 88.9/84.8 & 87.1/85.8 & 71.2/89.2 & 84.6/83.4 & 90.5 & 66.4  & 78.3 \\ \hline
        u-PMLM-A  & 56.5 & 94.3 & 88.8/84.4 & 87.0/85.9 & 71.4/89.2 & 84.5/83.5 & 91.8 & 66.1  & 79.0 \\ \hline
        u-PMLM-R  & 58.0 & 94.0 & 89.7/85.8 & 87.7/86.8 & 71.2/89.2 & 85.0/84.1 & 92.3 & 69.8  & 80.0 \\ \hline
        u-PMLM-R* & 56.9 & 94.2 & 90.7/87.7 & 89.7/89.1 & 72.2/89.4 & 86.1/85.4 & 92.1 & 78.5  & 81.3 \\ \hline
    \end{tabular}
    \caption{Evaluation on GLUE test set.}
    \label{glue}
\end{table*}
\begin{table}
\small
    \centering
    \begin{tabular}{|l|l|l|}
        \hline
        Model        & F1&EM \\ \hline
        BERT(A)    &76.85&73.97   \\ \hline
        u-PMLM-A  & 78.31&74.62  \\ \hline
        u-PMLM-R  &81.52&78.46  \\ \hline
    \end{tabular}
    \caption{Evaluation on SQUAD 2.0.}
    \label{squad}
\end{table}

Besides evaluating the ability of u-PMLM for natural language generation, we also evaluate its performance on natural language understanding. Two widely adopted tasks, GLUE~\cite{wang2018glue} and SQUAD 2.0~\cite{rajpurkar2018know}, are employed for evaluating u-PMLM. We have ensured that the evaluation for u-PMLM is influenced by as less model-irrelevant factors as possibles. For example, we do not tune the hyper-parameters and just follow the settings of BERT, including warming-up steps, learning rate, etc. In addition, since BERT employs absolute positional embeddings, the variant with absolute positional embeddings, u-PMLM-A, is intentionally trained for fairly evaluating the probabilistic masking scheme.

The results are shown in Table \ref{glue} and Table \ref{squad}. u-PMLM-A general performs better than BERT, demonstrating that the probabilistic masking scheme is more effective than the fixed masking scheme. The reason could be that the probabilistic masking scheme covers more a wider range of masking patterns, which benefits pretraining for a masked language model. Moreover, u-PMLM-R performs better than u-PMLM-A consistently. The only difference between these two models is the way to handle positional embedding. Relative positional embedding emphasizes more on the relative positions between two tokens, which could be a better option to capture contextual representation. Recall that relative and absolute positional embedding do not make many differences regarding generation ability if the dataset is proper. Hence we conclude u-PMLM-R is a better model than u-PMLM-A considering both NLU and NLG tasks. In addition, u-PMLM-R*, finetuned with a commonly used technique by sharing data from multiple tasks, is the state-of-the-art base model (with 110M parameters) trained on the BookCorpus and Wikipedia datasets on GLUE leaderboard on the date of paper submission.~\footnote{\url{http://gluebenchmark.com/leaderboard/}}

\paragraph{Comparison with XLNet}
\begin{table}[]
\small
    \centering
    \begin{tabular}{|l|l|l|l|}\hline
    Model &SQUAD 2.0 & MNLI& SST2 \\ 
    &F1/EM&m/mm& \\ \hline
XLNet (R) &81.33/78.46& 85.84/85.43 &92.66\\ \hline
u-PMLM-R &81.52/78.46 &85.99/85.60 &93.58 \\  \hline
    \end{tabular}
    \caption{Comparison with XLNet.}
    \label{tab:xlnet}
\end{table}
We also compare our proposed model with XLNet-base, which adopts relative positional embedding. As will be discussed in Section \ref{related-work}, XLNet is the most relevant model to u-PMLM. We are not able to train an XLNet using the same settings except that we make sure both u-PMLM-R and XLNet-base are of the same model size and are both trained using the same datasets. The comparison results shown in Table \ref{tab:xlnet} demonstrate that the performance of our proposed u-PMLM-R is comparable to XLNet.
\section{Related Work}\label{related-work}
\subsection{Non-traditional Text Generation}
Conventionally, text is commonly generated autoregressively in the left-to-right direction. Recently, some research works have proposed several models for non-autoregressive text generation \cite{welleck2019non,gu2019levenshtein}. \citet{insertion} proposes insertion Transformer, where text are generated in an iterative and partially autoregressive manner based on insertion operations. \citet{flowseq} design a latent variable based method to generate all the tokens in one pass. \citet{maskpredict} and \citet{bertmouth} employ masked language model for refinement-based non-autoregressive text generation, when a subset of tokens in a sequence are refined iteratively. Later, \citet{generalizedframework} propose a generalized framework of sequence generation accommodating autoregressive, semi-autoregressive, and refinement-based non-autoregressive model. Strictly speaking, our proposed arbitrarily ordered autoregressive text generation is a special case of this generalized framework. We are the first work to address such kind of text generation, which enables a lot of new applications over tradition text generation.

UNILM \cite{unilm} and MASS \cite{mass} are another two works that combine masked language model and autoregressive language model. However, UNILM only combines the training objective of GPT and BERT. MASS employs mask mechanism to train sequence to sequence language model. Both models do not address arbitrarily ordered text generation. 
\subsection{XLNet}
XLNet \cite{xlnet} is the most relevant pretrained language model to u-PMLM. Both of them can be treated as an autoregressive permutated language model.
However, XLNet is trained by permutating only a small fraction of the sequences, which does not fully address the generation problem. Though, we suppose that the training method for XLNet is feasible to train a model for arbitrarily ordered text generation as well. The main difference between these two models is that XLNet employs unidirectional Transformer, while u-PMLM is based on bidirectional Transformer. Regarding the training algorithm, XLNet shuffles the attention matrix and introduce two-stream self-attention, which is a bit complex and memory consuming. On the other hand, PMLM takes the simple training objective of masked language model and approximates permutated language model.
\section{Conclusion}
We have proposed a probabilistically masked language model for autoregressive generation in arbitrary word order. The experiments show that the text generated in arbitrary order has comparable quality with GPT. Besides, the proposed probabilistic masking scheme also improves the NLU capability of a masked language model.
\bibliography{acl2020}
\bibliographystyle{acl_natbib}
\appendix
\section{Proof of Equivalence} \label{proof}
We prove that PMLM with a continuous uniform distribution on the masking ratio, namely u-PMLM, is equivalent to an autoregressive permutated language model.

When $p(r)$ is a continuous uniform distribution, the probability $p(M)$ is analytical, denoted as:
\begin{equation}
\small
    \begin{split}
        p(M) &= \int r^K(1-r)^{N-K} p(r)d(r)\\
             &= \int_0^1 r^K(1-r)^{N-K}d(r)\\
                &= B(N-K+1,K+1) \\
           & = \frac{\Gamma (N-K+1)\Gamma (K+1)}{\Gamma (N+2)} \\
            & = \frac{(N-K)!K!}{(N+1)!} \\
    \end{split}
\end{equation}
\noindent where $B(\cdot)$ is Beta function and $\Gamma (\cdot)$ is Gamma function. Thus for u-PMLM, the expected loss-likelihood function denoted in Equation \ref{PMLM_X} becomes:
\begin{equation}\label{upmlm_1}
\small
\begin{split}
L_{\mathrm{pmlm}}(&X_{\Pi}|X;\theta) \\
=& \sum_M [\frac{1}{K}\sum_{k = 1}^K \log p(x_{\pi_k}|X_{-\Pi};\theta)]\frac{(N-K)!K!}{(N+1)!}\\
=& \frac{\sum_M \sum_{k = 1}^K (N-K)!(K-1)!\log p(x_{\pi_k}|X_{-\Pi};\theta)}{(N+1)!}
\end{split}
\end{equation}

On the other hand, we rewrite the formulation of an autoregressive permutated language model (APLM) denoted in Equation \ref{aplm} as:
\begin{equation} \label{aplm_2}
\small
\begin{split}
L_{\mathrm{aplm}}(X) &= \mathbb{E}_{\sigma} \left[\sum_{t=1}^{N}\log p(x_{\sigma_t}|x_{\sigma_1},\ldots,x_{\sigma_{t-1}};\theta) \right]\\
&= \frac{\sum_{\sigma} [\sum_{t=1}^{N}\log p(x_{\sigma_t}|x_{\sigma_1},\ldots,x_{\sigma_{t-1}};\theta) ]}{C}
\end{split}
\end{equation}
\noindent where the numerator sums over the log-likelihood for all the possible permutations and the denominator $C$ is a constant. In fact, we can rewrite the term $p(x_{\sigma_t}|x_{\sigma_1},\ldots,x_{\sigma_{t-1}};\theta)$ by $p(x_{\sigma_t}|X_{-\Pi^\sigma_{t}};\theta )$, where $\Pi^\sigma_{t} = X - \{\sigma_1, \sigma_2, ...,\sigma_{t-1}\}$. Noted that $K$ is the size of $\Pi^\sigma_{t}$. Thus the size of $\Pi^\sigma_{t}$ is denoted as $|\Pi^\sigma_{t}| = K = N-t+1$. Therefore we rewrite Equation \ref{aplm_2} as:
\begin{equation} \label{aplm_3}
\small
\begin{split}
&L_{\mathrm{aplm}}(X)\\
=& \frac{1}{C}\sum_{\sigma} [\log p(X_{\Pi^\sigma_{t+1}}|X_{-\Pi^\sigma_{t+1}};\theta )+\log p(x_{\sigma_{t}}|X_{-\Pi^\sigma_{t}};\theta )\\
&+\log p(X_{-\Pi^\sigma_{t}};\theta ) ]
\end{split}
\end{equation}
According to the above equation, we can derive the duplication factor for specific term $\log p(x_{\sigma_{t}}|X_{-\Pi^\sigma_{t}}) $ when summing over all the permutations, which is exactly the product of numbers of permutations for 
$\Pi^\sigma_{t+1}$ and $-\Pi^\sigma_{t}$ in the first and last term respectively. Specifically, the number of permutations for $\Pi^\sigma_{t+1}$ and $-\Pi^\sigma_{t}$ are factorials of $|\Pi^\sigma_{t+1}|$ and $|-\Pi^\sigma_{t}|$, denoted as:
\begin{equation}
\small
    \begin{split}
\mathrm{permutation}(\Pi^\sigma_{t+1}) = |\Pi^\sigma_{t+1}|! = (N-K)! \\
\mathrm{permutation}(-\Pi^\sigma_{t}) = |-\Pi^\sigma_{t}|! = (K-1)!
    \end{split}
\end{equation}
Hence the duplication factor for $\log p(x_{\sigma_{t}}|X_{-\Pi^\sigma_{t}}) $ is computed as $(N-K)!(K-1)!$, which is the coefficient of the expected log-likelihood function of u-PMLM denoted in Equation \ref{upmlm_1}. Thus we conclude that Equation \ref{upmlm_1} is equivalent to Equation \ref{aplm}, where the constant $C = (N+1)!$.
\section{Generation Examples of u-PMLM and BERT}\label{text-gen-upmlm-bert}
We show more examples of the text generated by u-PMLM-A and BERT respectively. Note that we do not manually select examples. Instead, these examples are picked randomly for fair comparison.

Below are texts generated by u-PMLM-A:
\textit{
\begin{enumerate}
    \item \textbf{Tom is a cat and Jerry is a mouse .} Though initially wary that he might steal the pie , Jerry eventually admits his fears . Tom ' s father , Selma , is upset when the cat is not present . She tries making a play about the pie but the girl ignores her and insists that she will be glad if he ' s able to steal the pie . \textbf{Tom and Jerry become good friends in the end .}
    \item \textbf{Tom is a cat and Jerry is a mouse .} `` Jerry and Tom meet again in a restaurant and in a confrontation Tom tries to call the house police . Jerry ' s neighbor Tom Martin ( Tom Lee ) finds them flirting by taking Jerry , but Tom finds out . With Jerry ' s help Tom gets their hand together in a car wash and the two have a date . \textbf{Tom and Jerry become good friends in the end .}
    \item \textbf{Tom is a cat and Jerry is a mouse .} When the kids get their first game of cricket at his home , he has him as catcher . When all of the teams are on the field except the school cricket player , he takes part in every game , which is always the cricket . He later breaks his leg on the last day of cricket and falls down a ladder . \textbf{Tom and Jerry become good friends in the end .}
    \item \textbf{Tom is a cat and Jerry is a mouse .} The two start off on a fishing trip together to see what their parents would think of them . The boys first discover a bad fish on the ocean and decide to try the game . A boathouse is built in town where they see a big picture of Tom ' s grandfather . The boys fight it out by chance , and \textbf{Tom and Jerry become good friends in the end .}
    \item \textbf{Tom is a cat and Jerry is a mouse .} This makes Tom very hard to understand why he and the other citizens like him , so he tries to have to get rid of the mouse . His best friend , Pete , tries to keep his dog quiet about this , even while Jerry and Marge insist they are just starting on a date due to Tom ' s dog . \textbf{Tom and Jerry become good friends in the end .}
\end{enumerate}
}

Below are texts generated by BERT:
\textit{
\begin{enumerate}
    \item  \textbf{Tom is a cat and Jerry is a mouse .} `` That same month the two characters were portrayed as their characters at the NBC series The Office , with the character played by Critrean Figgins and the mouse played by the cat . The two characters make a romantic connection and meet in the end of the episode " When We Were Meeted Together , " seeing why \textbf{Tom and Jerry become good friends in the end .}
    \item  \textbf{Tom is a cat and Jerry is a mouse .} . . And it ' s about when Tom is in trouble and he becomes frustrated when everything gets upsetting . I make my own movie . . . But I do not want to be a film so sad or about what I like about a kid ' s life . I want to make a movie that is more real and \textbf{Tom and Jerry become good friends in the end .}
    \item \textbf{Tom is a cat and Jerry is a mouse .} `` On August 7 , 1999 , DeLanto started a Twitter account , which included an online forum address on NBC ' s show 30 Rock through his account of his life on stage and on the Internet . During 2008 , he also posted on his personal blog a message saying " This world ' s really getting bigger . \textbf{Tom and Jerry become good friends in the end .}
    \item \textbf{Tom is a cat and Jerry is a mouse }. He is a cat and Tom is a mouse . At McKinley High School , Tom enters the Class 3A , and then is elected President of High School ( H . F . R . ) , the district ' s popular high school . He becomes the principal and a student ' s supervisor at the High School in 2004 . \textbf{Tom and Jerry become good friends in the end .}
    \item \textbf{Tom is a cat and Jerry is a mouse .} In April 1997 , Jack was murdered and he and Jack went on a similar out of wedlock . Tom eventually had a teenage son named Tim . In the pilot episode , Tom is shot in a car crash , and eventually re @ - @ takes his life after another accident , giving him a more " normal " appearance . \textbf{Tom and Jerry become good friends in the end .}
\end{enumerate}
}

\end{document}